\documentclass[letterpaper]{article} 
\usepackage{aaai2026}  
\usepackage{times}  
\usepackage{helvet}  
\usepackage{courier}  
\usepackage[hyphens]{url}  
\usepackage{graphicx} 
\usepackage{natbib}  
\usepackage{caption} 
\usepackage{algorithm}
\usepackage{algorithmic}

\usepackage{newfloat}
\usepackage{listings}
\DeclareCaptionStyle{ruled}{labelfont=normalfont,labelsep=colon,strut=off} 
\floatstyle{ruled}
\newfloat{listing}{tb}{lst}{}
\floatname{listing}{Listing}
\usepackage{tabularx}
\usepackage{pifont}
\usepackage{amsmath}
\usepackage{xspace}
\usepackage{multirow}
\usepackage{makecell}
\usepackage{booktabs}
\usepackage[table,xcdraw]{xcolor}
\newcommand*{\ourdataset}{\texttt{MAU-Set}}
\newcommand{\xmark}{\ding{55}}
      
\newcommand{\corrauthor}{\textsuperscript{\dag}}

\title{MAU-GPT: Enhancing Multi-type Industrial Anomaly Understanding via Anomaly-aware and  Generalist Experts Adaptation}
\author {
    Zhuonan Wang\textsuperscript{\rm 1}\equalcontrib,
    Zhenxuan Fan\textsuperscript{\rm 1}\equalcontrib,
    Siwen Tan\textsuperscript{\rm 1}\equalcontrib,
    Yu Zhong\textsuperscript{\rm 1},
    Yuqian Yuan\textsuperscript{\rm 1},
    Haoyuan Li\textsuperscript{\rm 1,2},
    Hao Jiang\textsuperscript{\rm 2},
    Wenqiao Zhang\textsuperscript{\rm 1}\corrauthor,
    Feifei Shao\textsuperscript{\rm 1}\corrauthor,
    Hongwei Wang\textsuperscript{\rm 1},
    Jun Xiao\textsuperscript{\rm 1}
}
\affiliations{
    \textsuperscript{\rm 1}Zhejiang University, \textsuperscript{\rm 2}Alibaba Group\\
    \{wnzz, zxfan, siwentan, yuzhong, yuqianyuan, lihaoyuan, wenqiaozhang, sff\}@zju.edu.cn\\
    aoshu.jh@alibaba-inc.com, hongweiwang@intl.zju.edu.cn, junx@cs.zju.edu.cn

}

\usepackage{bibentry}

\makeatletter

\newcounter{corrauthorfn}
\renewcommand{\corrauthor}{%
  \ifnum\value{corrauthorfn}=0
    \footnote{Corresponding authors.}%
    \setcounter{corrauthorfn}{\value{footnote}}%
  \else
    \footnotemark[\value{corrauthorfn}]%
  \fi
}

\makeatother

\begin{document}

\maketitle

\begin{abstract}
As industrial manufacturing scales, automating fine-grained product image analysis has become critical for quality control. However, existing approaches are hindered by limited dataset coverage and poor model generalization across diverse and complex anomaly patterns. To address these challenges, 
we introduce \ourdataset{}, a comprehensive dataset for \textbf{M}ulti-type industrial \textbf{A}nomaly \textbf{U}nderstanding. 
It spans multiple industrial domains and features a hierarchical task structure, ranging from binary classification to complex reasoning.
Alongside this dataset, we establish a rigorous evaluation protocol to facilitate fair and comprehensive model assessment.
Building upon this foundation, we further present \textbf{MAU-GPT}, a domain-adapted multimodal large model specifically designed for industrial anomaly understanding. It incorporates a novel AMoE-LoRA mechanism that unifies anomaly-aware and generalist experts adaptation, enhancing both understanding and reasoning across diverse defect classes. Extensive experiments show that MAU-GPT consistently outperforms prior state-of-the-art methods across all domains, demonstrating strong potential for scalable and automated industrial inspection. 
\end{abstract}



\begin{figure*}[t]
\centering
\includegraphics[width=\textwidth]{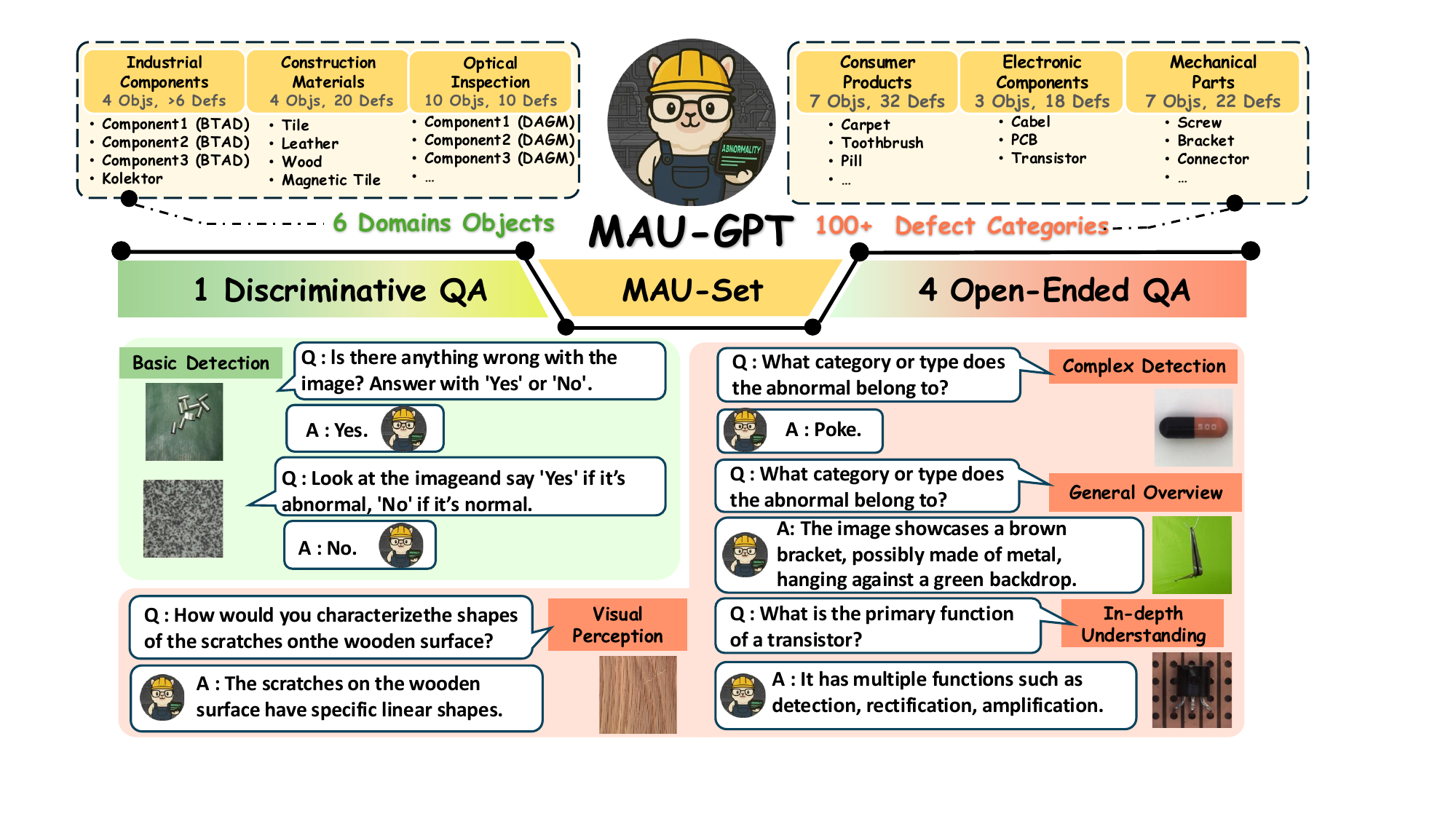}
\caption{Overview of the \textbf{MAU-GPT} model and \ourdataset{} dataset. The dataset spans \textbf{35} product types and over \textbf{100} defect categories across \textbf{6} major industrial domains, supporting \textbf{5} tasks ranging from discriminative question answering (QA) to open-ended visual reasoning. \textit{Objs} refer to object types, while \textit{Defs} denote defect categories. }
\label{fig:1}
\end{figure*}

\section{Introduction}

\addtolength{\tabcolsep}{0.5pt} 
\begin{table*}[t]
    \centering
    \scriptsize 
    \begin{tabular}{l|cccccccc} 
        \toprule
        & & & \multicolumn{3}{c}{\textbf{Image Number}} & & & \\
        \cline{4-6}
        \multirow{-2}{*}{\textbf{Dataset}} & \multirow{-2}{*}{\textbf{\makecell{Product \\ Type}}} & \multirow{-2}{*}{\textbf{\makecell{Defect \\ Class}}} & \textbf{Normal} & \textbf{Abnormal} & \textbf{All} & \multirow{-2}{*}{\textbf{\makecell{Annotation \\ Level}}} & \multirow{-2}{*}{\textbf{QA Type}} & \multirow{-2}{*}{\textbf{\makecell{QA Pairs \\ Constructed}}}\\
        \midrule
        BTAD \citep{mishra2021vt} & 3 & 3 & 2,250 & 580 & 2,830 & Segmentation mask & \xmark & 16,863 \\
        DAGM \citep{wieler2007weakly} & 10 & 10 & 10,000 & 1,500 & 11,500 & Segmentation mask & \xmark & 109,715 \\
        DeepPCB \citep{tang2019online} & 1 & 6 & 1,501 & 1,500 & 3,001 & Bounding box & \xmark & 21,090 \\
        KolektorSDD2 \citep{bovzivc2021mixed} & 1 & 3+ & 2,979 & 356 & 3,335 & Segmentation mask & \xmark & 22,274 \\
        Magnetic Tile \citep{huang2020surface} & 1 & 5 & 952 & 383 & 1,335 & Segmentation mask & \xmark & 9,002 \\
        MPDD \citep{jezek2021deep} & 4 & 8 & 1,064 & 282 & 1,346 & Segmentation mask & \xmark & 8,817 \\
        MVTec AD \citep{bergmann2019mvtec} & 15 & 73 & 4,096 & 1,258 & 5,354 & Segmentation mask & \xmark & 36,580 \\
        \midrule
        \rowcolor[HTML]{E9F3FE} \textbf{\texttt{MAU-Set} (Ours)} & \textbf{35} & \textbf{108+} & \textbf{22,842} & \textbf{5,859} & \textbf{28,701} & \textbf{Instance-level QA} & \textbf{Open/Close} & \textbf{224,341} \\
        \bottomrule
    \end{tabular}
    \caption{Comparison of widely used industrial anomaly detection benchmark datasets with our \texttt{MAU-Set}. QA-related columns (QA Type and QA Pairs Constructed) and the annotation level correspond to our newly constructed VQA annotations.}
    \label{data_source}
\end{table*}
\addtolength{\tabcolsep}{-0.5pt} 

Industrial anomaly analysis plays a vital role in product quality assessment, process monitoring, and defect prevention \citep{cao2024survey,ren2022state,bhatt2021image}.
In practice, quality inspectors conduct detailed quantitative evaluations and generate anomaly reports based on product images, relying on substantial domain expertise to identify subtle or complex defects \citep{tang2022industrial,Jiang2024mmad,baitieva2024supervised}. However, as production scales, manual inspection becomes increasingly impractical for delivering timely and consistent analysis \citep{alzarooni2025anomaly}, raising a critical question: \textit{Can we automate the in-depth analysis of industrial product images?}

Automating this process requires models capable of both fine-grained visual understanding and the reasoning of precise, context-aware descriptions. This challenge naturally aligns with methodologies in Visual Question Answering (VQA), where models are designed to perform image-grounded reasoning by interpreting visual content in response to textual queries and producing semantically rich answers \citep{chen2021finqa,lu2022learn, li2022devil}.

However, directly transferring existing VQA models from general-purpose domains to industrial anomaly analysis poses significant challenges. These challenges include, but are not limited to:
\textbf{1) Limited Dataset Coverage:} Existing public datasets are often narrowly scoped, focusing on specific anomaly types or product categories and failing to reflect the diversity of real-world industrial defects. For example, MVTec AD \citep{bergmann2019mvtec} primarily targets surface-level visual defects, while domain-specific datasets such as DeepPCB \citep{tang2019online} are restricted to anomalies in printed circuit boards.
\textbf{2) Poor Model Generalization:} The limited coverage of such datasets constrains model exposure to diverse defect patterns, impairing their ability to retrieve relevant domain knowledge. As a result, existing approaches struggle to differentiate fine-grained anomaly types and fail to accurately reason about their root causes.

To tackle the aforementioned dataset limitation, we construct \textbf{\texttt{MAU-Set}}, a comprehensive dataset for \textbf{\underline{M}}ulti-type industrial \textbf{\underline{A}}nomaly \textbf{\underline{U}}nderstanding.

MAU-Set defines two Question Answer (QA) styles---\textbf{\textit{Discriminative QA}} and \textbf{\textit{Open-Ended QA}}---covering tasks from basic binary classification to complex reasoning, and further groups them into five distinct tasks with varying difficulty levels.
This hierarchical task structure increases both the complexity and depth of the benchmark, promoting progressive knowledge accumulation and equipping models to handle the varied challenges encountered in real-world industrial applications.
As illustrated in Figure~\ref{fig:1} and Table~\ref{data_source}, MAU-Set spans \textbf{35 product types and over 100 defect classes across 6 industrial domains}, ensuring both breadth and diversity. 
In addition, 
MAU-Set provides broad anomaly coverage, fine-grained supervision, and a unified testbed, laying a solid foundation for the development and evaluation of scalable, automated, and trustworthy industrial anomaly analysis systems.

To address the challenges of limited generalization and restricted reasoning capacity, we propose \textbf{MAU-GPT}, a multimodal large model specifically tailored for industrial anomaly understanding. Built upon a pre-trained vision-language foundation, MAU-GPT introduces a novel Adaptive Mixture of Experts with Low-Rank Adaptation (\textbf{AMoE-LoRA}) mechanism that unifies anomaly-aware and generalist experts adaptation, enhancing both generalization and reasoning across diverse anomaly classes. Concretely, for general samples, MAU-GPT draws on the Mixture-of-Experts (MoE) paradigm by dynamically activating multiple general LoRA experts via input-dependent routing. This flexible expert selection alleviates the capacity limitations of single-LoRA modules and enables fine-grained, sample-aware representation learning. For anomalous samples, we integrate an anomaly-aware adaptation strategy in which a hypernetwork generates LoRA parameters on-the-fly, conditioned on the specific anomaly category. This eliminates the need for pre-trained category-specific LoRA modules and allows the model to adapt to previously unseen defects. 
Crucially, this hybrid adaptation framework facilitates seamless model generalization across diverse domains while enhancing sensitivity to rare and novel anomalies, enabling accurate detection in dynamic and complex industrial environments.

The contributions of this work are summarized as follows:

\begin{itemize} 

\item \textbf{A Comprehensive Dataset.} We present MAU-Set, a broad-coverage, fine-grained dataset that introduces two QA
styles---\textit{Discriminative QA} and \textit{Open-Ended QA}---designed to evaluate both low-level perception and high-level visual reasoning across diverse industrial domains.

\item \textbf{A Domain-Adaptive Model.} We propose MAU-GPT, a multimodal large model equipped with a novel AMoE-LoRA mechanism, enabling accurate and robust anomaly understanding across diverse domains. 
Extensive experiments show its strong potential and generalization capabilities for scalable and automated industrial inspection.

\item \textbf{Open-Source Work.} We will open-source the dataset, model checkpoints, and codebase to facilitate further research and development in industrial anomaly analysis.

\end{itemize}

\section{Related Work}
\noindent\textbf{$\bullet$ Industrial Anomaly Detection.} Industrial anomaly detection methods are typically divided into unsupervised and supervised approaches. Unsupervised methods do not rely on labeled data, and are mainly based on feature embedding or reconstruction. Feature embedding methods, such as SVDD \citep{tax2004support} and PatchSVDD \citep{yi2020patch}, distinguish normal and abnormal data via hypersphere-based or probabilistic modeling. Reconstruction-based approaches like Autoencoders \citep{bergmann2018improving} and GANs \citep{yan2021learning} use reconstruction loss to identify anomalies. Recently, transformer-based models \citep{mishra2021vt} and diffusion methods \citep{wyatt2022anoddpm} have shown promising results in industrial settings.
In contrast, supervised methods utilize limited labeled anomalies. Some adopt semi-supervised learning \citep{chu2020neural} or loss-based training \citep{liznerski2020explainable}, while others focus on fine-grained anomaly modeling\citep{pang2021explainable, shao2024knowledge} or disentangling anomaly representations \citep{ding2022catching,shao2021improving}. Although these approaches depend on annotations, they often alleviate data imbalance by introducing pseudo-labels or synthesizing anomalies.

\noindent\textbf{$\bullet$ Large Vision-Language Models.} 
The rapid advancements in large language models (LLMs) have spurred the development of Large Vision-Language Models (LVLMs), which integrate powerful linguistic understanding with visual perception. These models, often built upon frozen image encoders and large pre-trained LLMs, demonstrate remarkable capabilities in multimodal understanding, reasoning, and generation across diverse tasks such as visual question answering, image captioning, and visual dialogue ~\citep{alayrac2022flamingo,li2023blip, li2023zero}. By leveraging extensive pre-training on vast amounts of image-text pairs, LVLMs learn rich cross-modal representations, enabling them to comprehend complex visual scenes and respond to intricate textual queries. Recent architectures, such as LLaVA \citep{liu2024visual}, have notably advanced visual instruction tuning, while InternVL \citep{chen2024expanding} has demonstrated significant progress in scaling up vision foundation models for large-scale visual document understanding, further refining efficiency and performance, and allowing for more versatile applications. The emergence of these sophisticated LVLMs presents a promising avenue for addressing complex visual analysis challenges in specialized domains, including industrial anomaly detection, 

\section{Dataset: \texttt{MAU-Set}}
\label{sec:dataset}
\begin{figure}[t]
    \centering
    \includegraphics[width=\linewidth]{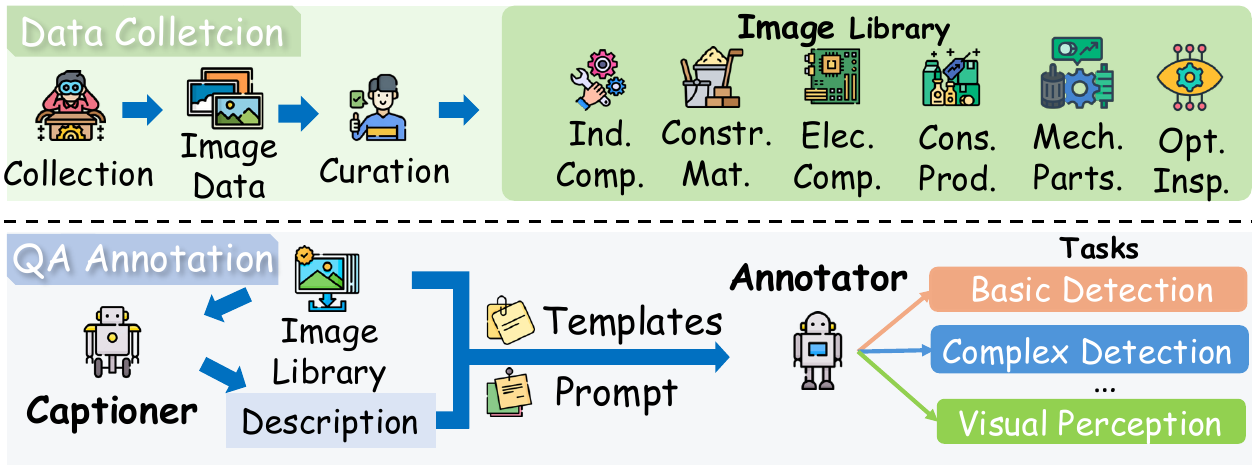}
     \caption{The data collection and annotation of \ourdataset{}.
     \textbf{Ind. Comp.} and \textbf{Constr. Mat.} denote Industrial Components and Construction Materials; 
      \textbf{Elec. Comp.,Cons. Prod.,Mech. Parts} and \textbf{Opt. Insp.} are defined analogously.}
    \label{dataset_process}
\end{figure}

To support fair and comprehensive evaluation in industrial anomaly analysis, we introduce \ourdataset{}, which features six industrial domains, two progressive QA-style questions, ranging from basic binary classification to complex anomaly reasoning, with detailed collection and annotation provided in Figure \ref{dataset_process}. 
This hierarchical task structure increases both the complexity and depth of the benchmark, encouraging models to progressively accumulate knowledge and adapt to the diverse challenges encountered in practical scenarios.


\subsection{Dataset Design}

\begin{figure}[t]
    \centering
    \includegraphics[width=0.98\linewidth]{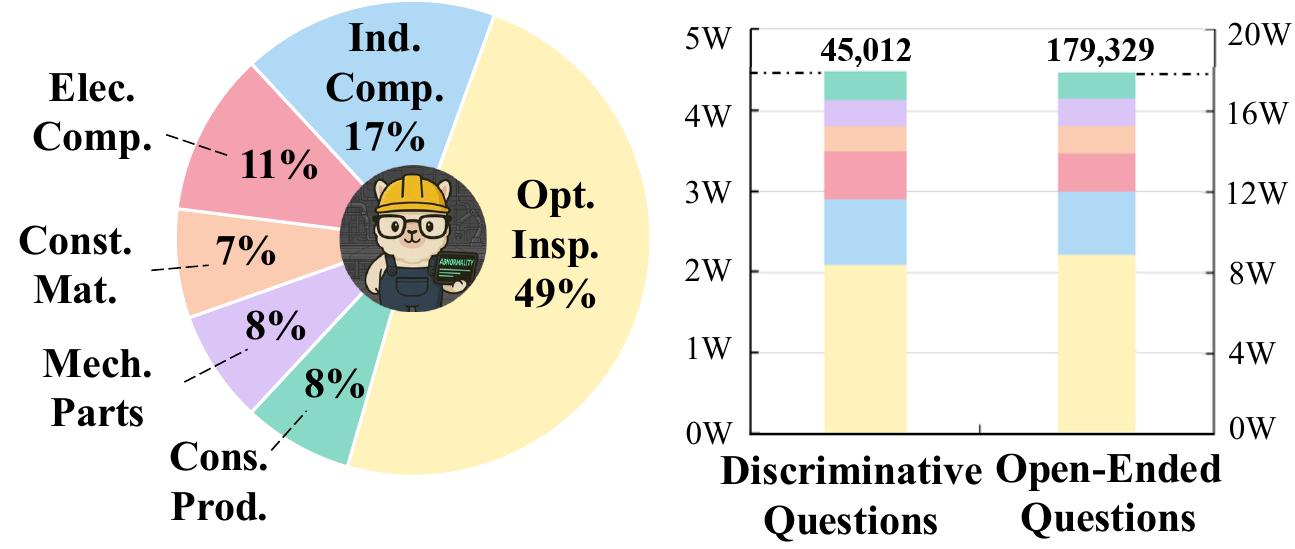}
    \caption{The data collection and annotation of \ourdataset{}.}
    \label{fig:data_statistics}
\end{figure}

\begin{itemize}
\item \noindent\textbf{Data Collection.}
As illustrated in Figure~\ref{dataset_process}, we construct the dataset by curating 7 data sources, incorporating both real and synthetic images to enable robust evaluation and foster progress in industrial anomaly analysis. As shown in Table~\ref{data_source}, the dataset comprises 35 product types, encompassing both surface textures and objects themselves. 

\item \noindent\textbf{Domain Division.} To enhance field clarity, we further organize these sources into 6 industrial domains. As reported in Figure~\ref{fig:data_statistics}, these domains captures the breadth and complexity of real-world industrial scenarios. 

\item \noindent\textbf{QA Style.}
To fully utilize existing data and support diverse evaluation, we design two QA styles:
(i) Discriminative QA: Following the conventional detection formulation, the model determines whether an image contains an anomaly and produces a binary output: \textit{True} (abnormal) or \textit{False} (normal).
(ii) Open-Ended QA: It requires the model to perform fine-grained reasoning by generating free-form responses to image-specific questions. These questions may involve identifying defect types, localizing defects, or inferring causes and implications, etc.

\item \noindent\textbf{Task Division.}
To improve the clarity of the task structure, we further decompose the two QA types into 5 distinct tasks. As illustrated in Figure 1, Discriminative QA corresponds to Basic Detection, while Open-Ended QA is subdivided into 4 tasks of increasing difficulty. 
\end{itemize}

\subsection{Data Annotation}
We collect plenty of images from 7 sources, however, some sources are designed for (semi-)unsupervised training, resulting in significant class imbalance between normal and abnormal samples in both training and test sets. 
To address this, we first collect normal and abnormal samples from each source, apply a fixed-ratio split for training and testing, and then merge them to create a more balanced dataset.

Inspired by prior work on the benefits of instruction diversity, we build a question pool with varied prompts for each task. For each image, 1–2 questions per task are sampled, yielding around 7–8 questions across 5 tasks. Grammatically incorrect questions are manually filtered. Unlike other tasks, the ``In-depth Understanding'' task necessitates domain-specific expertise in industrial products. To meet this requirement, we manually curate a set of high-quality, expert-level questions tailored to the unique characteristics of each product. These questions are further used as prompts for a pre-trained large language model to automatically generate additional professional-grade questions, thereby streamlining the annotation process.

\subsection{Dataset Statistics}
As shown in Table~\ref{data_source} and Figure \ref{fig:data_statistics}, \ourdataset{} comprises 35 product types, over 100 defect classes, and approximately 28K images accompanied by 224K high-quality, diverse QA instruction pairs. The dataset spans 6 industrial domains and includes 2 styles of QA questions, which are further divided into 5 distinct tasks.
This design ensures broad coverage of real-world anomaly scenarios and promotes the domain adaptability of detection models. Additionally, its hierarchical task structure---from Discriminative QA (binary classification) to Open-Ended QA (complex reasoning)---facilitates progressive learning, enhancing model generalization and robustness in anomaly detection.

\begin{figure*}[t]
    \centering
    \includegraphics[width=\linewidth]{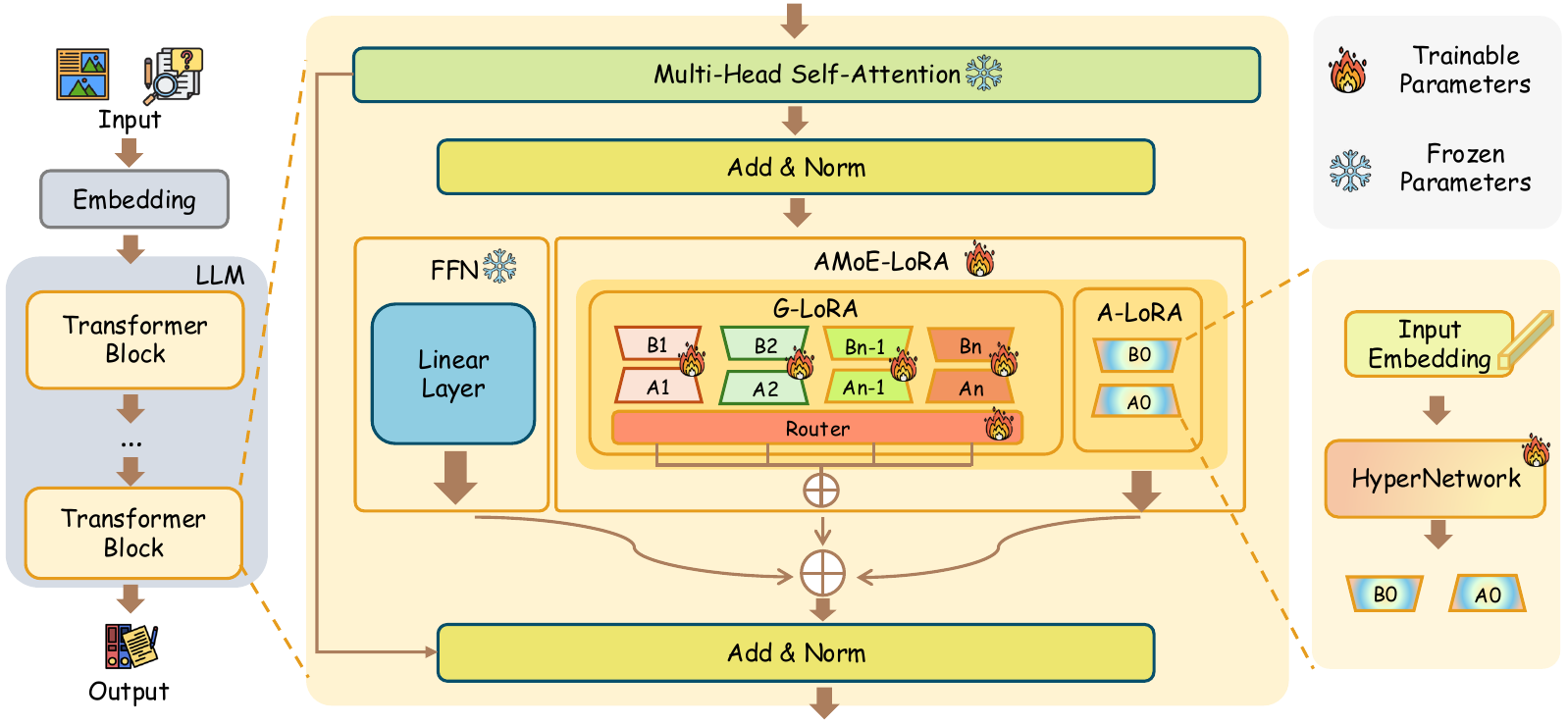}
    \caption{The AMoE-LoRA architecture combines generalist experts with an anomaly-aware expert, enabling the MLLM to incorporate both general knowledge and domain-specific insights for industrial anomaly understanding.}
    \label{fig:AMoE-LoRA}
    \vspace{-3mm}
\end{figure*}

\section{Method: MAU-GPT}
\label{sec:method}

\subsection{Overview}
As illustrated in Figure~\ref{fig:AMoE-LoRA}, MAU-GPT comprises a frozen vision encoder, a trainable visual projection layer, a word embedding layer, and a stack of LLM blocks. At the core of these blocks lies our proposed \textbf{AMoE-LoRA} module, which serves as the primary mechanism for domain adaptation and knowledge specialization.

This design allows MAU-GPT to leverage large models’ pre-trained knowledge and generalization while enabling parameter-efficient domain adaptation and specialization for diverse industrial domains. The AMoE-LoRA architecture and training procedure are described in Sections 4.2 and 4.3, respectively.

\subsection{Architecture of AMoE-LoRA}
AMoE-LoRA is a hybrid expert architecture that integrates both anomaly-aware and generalist experts adaptations. The outputs from these two expert branches are fused into the model by adding them to the original output. In this work, multiple AMoE-LoRA module are inserted at multiple transformer blocks within a multimodal large language model, respectively. For clarity, we will describe the AMoE-LoRA module from the perspective of a single transformer block.

Specifically, the generalist experts component adopts the MoE paradigm, enabling multiple experts (G-LoRA) to collaborate through input-dependent routing that dynamically adjust the contribution weight of each expert. This module comprises $N$ experts and a router. The output of the generalist experts is computed as follows:
\begin{equation}
o_1=\Delta W x=\sum_{i=1}^N \mathcal{R}(x)_i \cdot E_i(x), 
\end{equation}
where $\mathcal{R}(\cdot)=\operatorname{Softmax}\left(x W_g\right)$ denotes the routing function with router weights $W_g$, $x$ is the embedding of the original model, and $E_i(\cdot)$ represents the $i$-th expert. Each expert is parameterized by a pair of low-rank matrices $B_i$ and $A_i$ as follows:
\begin{equation}
E_i(x)=B_iA_ix.    
\end{equation}
Both the router weights $W_g$ and the low-rank matrices $B_i$ and $A_i$ are trainable. During forward propagation, the routing function $\mathcal{R}(\cdot)$ produces a contribution weight $\omega_i$ for each expert, effectively integrating knowledge from all experts. Thus, the output can be expressed as:
\begin{equation}
o_1=\frac{\alpha}{r} \sum_{i=1}^N \omega_i \cdot B_i A_i x,
\end{equation}
where $r$ is the rank of the matrices and $\alpha$ is the scaling factor that modulates the impact of the generalist experts on the overall model.

In addition, the anomaly-aware expert component leverages a trainable hypernetwork to generate adaptive LoRA (A-LoRA) weights, enabling the model to adapt to each individual sample. This eliminates the need for pre-trained category-specific LoRA modules and allows the model to adapt to previously unseen defects. Specifically, the embedding $x$ of the original model is fed into the hypernetwork $H$ to produce the parameters of a pair of low-rank matrices $A_0$ and $B_0$ as follows:
\begin{equation}
\begin{aligned}
& A_0 = W_aH(x), \\
& B_0 = H(x)W_b,
\end{aligned}
\end{equation}
where $W_a$ and $W_b$ are two learnable weight matrices. The output of the anomaly-aware expert is then defined as:
\begin{equation}
o_2=\frac{\alpha}{r}  \cdot B_0 A_0 x.
\end{equation}

Finally, the outputs of both generalist experts component and anomaly-aware expert component are added to the original model output $o_0$:
\begin{equation}
o=o_0+o_1+o_2,
\end{equation}
where $o$ denotes the final output.

\subsection{Training Pipeline}
In this work, we adopt a stage-wise training strategy, as suggested by previous studies~\cite{moryossef2019step, bolukbasi2017adaptive}, to simplify the learning process, enhance robustness, and improve manageability. This strategy facilitates progressive knowledge transfer, efficient parameter tuning, and effective specialization, enabling the model to retain general reasoning capabilities while adapting to complex industrial anomaly scenarios.

\noindent\textbf{$\bullet$ Stage 1: Multimodal Pre-training.}
The vision encoder and the LLM are kept frozen, while only the multimodal projection layer is trained. By Leveraging the high-quality instruction dataset, the model learns to align visual and textual modalities, thereby improving its ability to interpret and reason over image content.

\noindent\textbf{$\bullet$ Stage 2: Industrial Anomaly Instruction Fine-tuning.}
Building on the established vision-language alignment and general reasoning capabilities, we fine-tune the proposed AMoE-LoRA module and the projection layer to specialize the model for industrial anomaly detection.

\section{Experiments}
\label{sec:Experiments}
\subsection{Experimental Setup}
\noindent\textbf{$\bullet$ Data Details.} 
The first-stage training follows the same procedure as LLaVA, where we pretrain the model on the LLaVA-558K dataset \cite{liu2024improved}. In the second stage, we specialize the model for industrial anomaly detection by fine-tuning it on our proposed \ourdataset{} dataset. 

\noindent\textbf{$\bullet$ Model Settings.}
To comprehensively evaluate the performance of MAU-GPT, we compare it with a range of state-of-the-art models, including six mainstream open-source multimodal large language models (MMLMs), such as LLaVA-1.5 (with two parameter variants), Unified-IO2 (with two parameter variants), Gemma-3, Yi-VL, and InternVL-2.5. Additionally, we include AnomalyGPT, an MLLM specialized for industrial anomaly detection, which is evaluated in few-shot settings ranging from one-shot to four-shot.

\begin{table*}[ht]
  \centering
  
  \resizebox{\textwidth}{!}{
  \begin{tabular}{c|l|cccccc|c}
    \toprule
    \rowcolor{black!10}\textbf{Type} & \textbf{Model} & \makecell{\textbf{Industrial}\\ \textbf{Components}} & \makecell{\textbf{Electronic}\\ \textbf{Components}} & \makecell{\textbf{Construction}\\ \textbf{Materials}} & \makecell{\textbf{Mechanical}\\ \textbf{Parts}} & \makecell{\textbf{Consumer}\\ \textbf{Products}} & \makecell{\textbf{Optical}\\ \textbf{Inspection}} & \textbf{Avg.} \\
    \midrule
    \multirow{7}{*}{\textbf{Generalist Model}}
    & LLaVA-1.5(7B) & 40.10 & 30.21 & 39.46 & 50.66 & 45.83 & 55.40 & 43.61 \\
    & LLaVA-1.5(13B) & 11.57 & 27.31 & 25.26 & 32.13 & 27.76 & 15.24 & 23.21 \\
    & UIO2-Large(1B) & 66.61 & 53.26 & 53.28 & 57.06 & 61.61 & 66.40 & 59.70 \\
    & UIO2-Xl(3B) & 55.55 & 55.17 & 64.04 & 53.28 & 60.38 & 67.84 & 59.38 \\
    & Gemma-3(4B) & 63.06 & 58.12 & 43.75 & 48.64 & 51.50 & 33.86 & 42.45 \\
    & Yi-VL(6B) &  62.39 & 60.05 & 58.54 & 47.96 & 52.91 & 83.17 & 73.54 \\
    & InternVL-2.5(8B) & 46.58 & 57.77 & 47.71 & 49.66 & 55.31 & 41.62 & 45.31 \\
    \midrule
    \multirow{5}{*}{\textbf{Specialized Model}}
    & AnomalyGPT(1-shot) & 50.93 & 66.07 & 78.70 & 60.41 & \textbf{79.31} & 39.76 & 62.53 \\
    & AnomalyGPT(2-shot) & 56.20 & 70.16 & 79.46 & 64.38 & 75.79 & 41.25 & 64.54 \\
    & AnomalyGPT(3-shot) & 57.04 & 69.74 & 80.19 & 67.14 & 76.62 & 42.92 & 65.61 \\
    & AnomalyGPT(4-shot) & 58.14 &\textbf{ 69.78} & \textbf{81.00} & 66.44 & 77.41 & 43.39 & 66.03 \\
    &  \cellcolor[HTML]{E9F3FE}\textbf{MAU-GPT(4B)} & \cellcolor[HTML]{E9F3FE}\textbf{89.37} &\cellcolor[HTML]{E9F3FE} 53.73&\cellcolor[HTML]{E9F3FE}73.75& \cellcolor[HTML]{E9F3FE}\textbf{77.56} & \cellcolor[HTML]{E9F3FE}74.75 & \cellcolor[HTML]{E9F3FE}\textbf{86.81} & \cellcolor[HTML]{E9F3FE}\textbf{82.12}\\
    \bottomrule
  \end{tabular}
  }
  \caption{Performance of different models on Discriminative QA.}
  \label{Discriminative_qa}
\end{table*}

\begin{table*}[ht]
  \centering
  
  \resizebox{\textwidth}{!}{
  \begin{tabular}{c|l|cccccc|c}
    \toprule
    \rowcolor{black!10}\textbf{Type} & \textbf{Model} & \makecell{\textbf{Industrial}\\ \textbf{Components}} & \makecell{\textbf{Electronic}\\ \textbf{Components}} & \makecell{\textbf{Construction}\\ \textbf{Materials}} & \makecell{\textbf{Mechanical}\\ \textbf{Parts}} & \makecell{\textbf{Consumer}\\ \textbf{Products}} & \makecell{\textbf{Optical}\\ \textbf{Inspection}} & \textbf{Avg.} \\
    \midrule
    \multirow{7}{*}{\textbf{Generalist Model}}
    & LLaVA-1.5(7B) & 16.11 & 30.32 & 39.85 & 30.72 & 40.48 & 40.60 & 33.01 \\
    & LLaVA-1.5(13B) & 18.93 & 34.80 & 32.79 & 29.94 & 41.50 & 41.37 & 33.22 \\
    & UIO2-Large(1B) & 14.33 & 25.06 & 33.38 & 29.14 & 34.45 & 13.38 & 24.96 \\
    & UIO2-Xl(3B) & 21.29 & 26.60 & 33.92 & 25.39 & 34.02 & 16.90 & 26.35 \\
    & Gemma-3(4B) & 31.96 & 54.31 & 40.36 & 39.80 & 50.59 & 42.41 & 43.24 \\
    & Yi-VL(6B) & 31.47 & 39.73 & 34.96 & 40.89 & 47.84 & 36.59 & 38.58 \\
    & InternVL-2.5(8B) & 30.52 & 50.94 & 50.23 & 45.35 & 51.12 & 61.08 & 48.21 \\
    \midrule
    \multirow{5}{*}{\textbf{Specialized Model}}
    & AnomalyGPT(1-shot) & 11.19 & 29.27 & 25.12 & 17.97 & 37.08 & 9.00 & 21.61 \\
    & AnomalyGPT(2-shot) & 11.54 & 31.27 & 26.73 & 19.27 & 36.47 & 9.05 & 22.39 \\
    & AnomalyGPT(3-shot) & 12.00 & 29.30 & 27.01 & 19.83 & 36.54 & 9.23 & 22.32 \\
    & AnomalyGPT(4-shot) & 12.64 & 28.16 & 27.31 & 19.73 & 36.71 & 9.23 & 22.30 \\
    & \cellcolor[HTML]{E9F3FE}\textbf{MAU-GPT(4B)} & \cellcolor[HTML]{E9F3FE}\textbf{88.75} & \cellcolor[HTML]{E9F3FE}\textbf{81.66} & \cellcolor[HTML]{E9F3FE}\textbf{85.64} & \cellcolor[HTML]{E9F3FE}\textbf{84.37} & \cellcolor[HTML]{E9F3FE}\textbf{86.97} & \cellcolor[HTML]{E9F3FE}\textbf{94.39} & \cellcolor[HTML]{E9F3FE}\textbf{91.05} \\
    \bottomrule
  \end{tabular}
  }
  \caption{Performance of different models on GPT-4o judged accuracy for Open-Ended QA.}
  \vspace{-3mm}
  \label{gpt4-judge}
\end{table*}

\noindent\textbf{$\bullet$ Evaluation Metrics.}
To better evaluate the model’s performance, we adopt both automatic and GPT-4o-judged metrics. Specifically, we report ROUGE (ROUGE-1/2/L) \cite{ganesan2018rouge} and BLEU \cite{saadany2021bleu} to measure n-gram/sequence overlap between generated outputs and references. In addition, we leverage GPT-4o as an evaluator to assess answer correctness against the annotated ground truth, enabling a finer-grained evaluation.

\subsection{Main Experiments} 
\label{sec:eva}
\noindent\textbf{$\bullet$ Performance on Discriminative QA.}
Table~\ref{Discriminative_qa} reports results on the Discriminative QA task of \ourdataset{}. Early open-source models (e.g., LLaVA, UIO2) provide moderate baselines, and simply scaling (e.g., LLaVA-1.5) brings limited gains without domain alignment. More recent general-purpose VLMs (e.g., Yi-VL, InternVL-2.5) generally perform better, with Yi-VL being the most consistent.  Although AnomalyGPT benefits from more in-context examples, it still lags behind due to prompt-only specialization. Overall, MAU-GPT achieves the best performance, demonstrating the advantage of domain-specific fine-tuning.
\addtolength{\tabcolsep}{-1pt} 
\begin{table}[t]
  \centering
  \resizebox{0.48\textwidth}{!}{
  \begin{tabular}{l|cccc}
    \toprule
    \rowcolor{black!10}\textbf{Model} & \textbf{ROUGE-1} & \textbf{ROUGE-2} & \textbf{ROUGE-L} & \textbf{BLEU-4} \\
    \midrule
    
     LLaVA-1.5(7B) & 0.2495 & 0.0894 & 0.2008 & 0.0343 \\
     LLaVA-1.5(13B) & 0.2459 & 0.0884 & 0.1976 & 0.0332 \\
     UIO2-Large(1B) & 0.1312 & 0.0204 & 0.1284 & 0.0060 \\
     UIO2-Xl(3B) & 0.1666 & 0.0348 & 0.1556 & 0.0102 \\
    Gemma-3(4B) & 0.2936 & 0.0905 & 0.2425 & 0.0357 \\
     Yi-VL(6B) & 0.3044 & 0.1045 & 0.2469 & 0.0414 \\    
     InternVL-2.5(8B) & 0.2982 & 0.1103 & 0.2367 & 0.0440 \\
    \midrule
     AnomalyGPT(1-shot) & 0.1947 & 0.0516 & 0.1730 & 0.0216 \\
     AnomalyGPT(2-shot) & 0.1937 & 0.0514 & 0.1720 & 0.0217 \\
     AnomalyGPT(3-shot) & 0.1943 & 0.0519 & 0.1725 & 0.0216 \\
     AnomalyGPT(4-shot) & 0.1954 & 0.0521 & 0.1734 & 0.0216 \\
     \cellcolor[HTML]{E9F3FE}\textbf{MAU-GPT(4B)} & \cellcolor[HTML]{E9F3FE}\textbf{0.7026} & \cellcolor[HTML]{E9F3FE}\textbf{0.4537} & \cellcolor[HTML]{E9F3FE}\textbf{0.6531} & \cellcolor[HTML]{E9F3FE}\textbf{0.4773} \\
    \bottomrule
  \end{tabular}
  }
  \caption{Performance of different models on ROUGE and BLEU metrics for Open-Ended QA.}
  \vspace{-3mm}
  \label{open-end_qa}
\end{table}
\addtolength{\tabcolsep}{1pt}

\begin{table*}[!h]
  \centering
  \footnotesize
  \renewcommand{\arraystretch}{1.05}
  \setlength{\tabcolsep}{4pt} 
  \begin{tabularx}{\textwidth}{l|c|*{4}{>{\centering\arraybackslash}X}|*{2}{>{\centering\arraybackslash}X}|c}
    \toprule

    \rowcolor{black!10}
     & \textbf{Anomaly} &
    \multicolumn{4}{c|}{\textbf{Defect}} &
    \multicolumn{2}{c|}{\textbf{Object}} &
     \\
    \rowcolor{black!10}
    \multirow{-2}{*}{\textbf{Model}} & \textbf{Discrimination} &
    \textbf{Classification} & \textbf{Localization} & \textbf{Description} & \textbf{Analysis} &
    \textbf{Classification} & \textbf{Analysis} & \multirow{-2}{*}{\textbf{Average}} \\

    \midrule
    Random Chance       & 50.00 & 25.00 & 25.00 & 25.00 & 25.00 & 25.00 & 25.00 & 28.57 \\
    \midrule
    Human (expert)      & 95.24 & 75.00 & 92.31 & 83.33 & 94.20 & 86.11 & 80.37 & 86.65 \\
    \midrule
    GPT-4o-mini         & 64.33 & 48.58 & 38.75 & 63.68 & 80.40 & 88.56 & 79.74 & 66.29 \\
    GPT-4o              & \textbf{68.63} & \textbf{65.80} & \textbf{55.62} & \textbf{73.21} & \textbf{83.41} & \textbf{94.98} & \textbf{82.80} & \textbf{74.92} \\
    \midrule
    Uio2-large-1.5B     & 49.27 & 27.52 & 18.95 & 25.28 & 37.85 & 43.75 & 47.24 & 36.08 \\
    Uio2-Xl-6B          & 24.52 & 32.23 & 36.61 & 36.99 & 48.64 & 57.85 & 59.21 & 38.62 \\
    AnomalyGPT-7B       & 61.53 & 31.39 & 27.20 & 30.76 & 22.24 & 44.48 & 30.23 & 35.41 \\
    Gemma3-4B           & 57.37 & 37.07 & 35.71 & 47.49 & 76.32 & 85.11 & 73.65 & 58.96 \\
    Qwen-VL-Chat-7B     & 53.65 & 31.33 & 28.62 & 41.66 & 63.99 & 74.46 & 67.94 & 51.66 \\
    SPHINX-7B           & 53.13 & 33.93 & 52.27 & 50.96 & 71.23 & 85.07 & 73.10 & 59.96 \\
    LLaVA-1.5-13B       & 49.96 & 38.78 & 46.17 & 58.17 & 73.09 & 73.62 & 70.98 & 58.68 \\
    InternVL2-34B       & 59.97 & 43.85 & 47.91 & 57.60 & 78.10 & 74.18 & 80.37 & 63.14 \\
    LLaVA-NeXT-34B      & 57.92 & 48.79 & 52.87 & 71.34 & 80.28 & 81.12 & 77.80 & 67.16 \\
    
    \cellcolor[HTML]{E9F3FE}\textbf{MAU-GPT-4B} & \cellcolor[HTML]{E9F3FE}67.50 & \cellcolor[HTML]{E9F3FE}39.31 & \cellcolor[HTML]{E9F3FE}41.24 & \cellcolor[HTML]{E9F3FE}66.37 & \cellcolor[HTML]{E9F3FE}77.04 & \cellcolor[HTML]{E9F3FE}67.09 & \cellcolor[HTML]{E9F3FE}71.31 & \cellcolor[HTML]{E9F3FE}61.41 \\
    \bottomrule
  \end{tabularx}
  \caption{Accuracy of various models on the MMAD Benchmark across different industrial anomaly understanding tasks.}
  \label{tab:mmad}
\end{table*}

\begin{figure*}[t]
    \centering
    \vspace{-3mm}
    \includegraphics[width=1\linewidth]{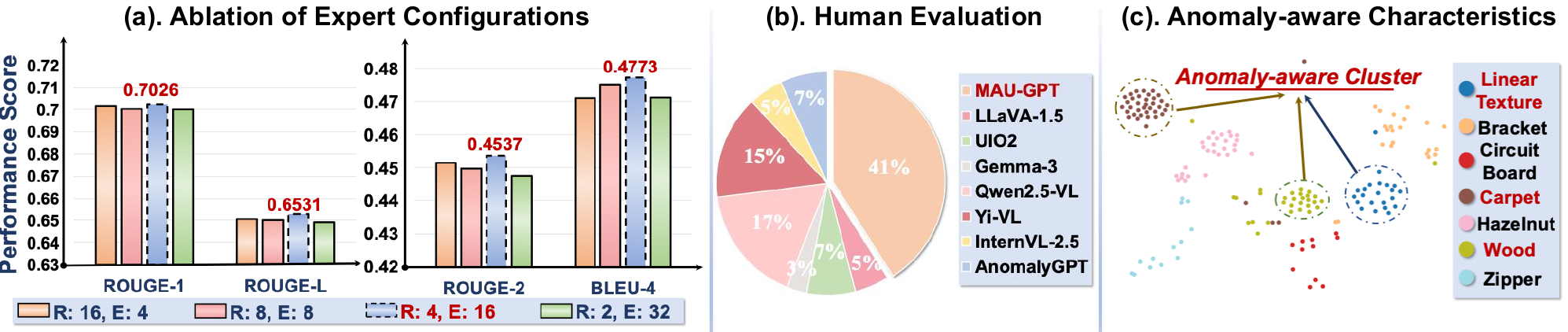}
     \vspace{-3mm}
    \caption{(a) describes the influence of the number and rank of the generalist experts. (b) shows that our model receives the highest endorsement from human experts. (c) is the 2D representation of hypernetwork-generated parameters across samples.}
    \label{fig:AMoE-LoRA_result}
    \vspace{-3mm}
\end{figure*}

\noindent\textbf{$\bullet$ Performance on Open-Ended QA.}
Compared to Discriminative questions, Open-Ended QA is inherently more challenging as models must generate coherent, contextually grounded text. Using GPT-4o as a no-reference evaluator alongside ROUGE and BLEU, we observe that newer general-purpose VLMs (e.g., InternVL, Gemma-3) outperform earlier baselines such as LLaVA and UIO2, while AnomalyGPT shows only limited gains with additional in-context examples—highlighting the limitations of prompt-only adaptation. In contrast, MAU-GPT achieves the strongest performance across all evaluation metrics, demonstrating that domain-tailored fine-tuning is crucial for producing accurate and semantically aligned open-ended responses in industrial anomaly reasoning.

\noindent\textbf{$\bullet$ Performance on MMAD Benchmark.}


Table~\ref{tab:mmad} summarizes results on MMAD, spanning seven tasks in anomaly discrimination and defect/object-level understanding. Although large general-purpose models (e.g., GPT-4o, LLaVA-NeXT) achieve higher absolute scores, they require substantially more parameters. Under comparable model sizes, MAU-GPT attains the best overall performance among open-source VLMs and approaches the accuracy of much larger 34B models, highlighting strong parameter efficiency for industrial anomaly understanding.

\subsection{Ablation and In-depth Study}
\noindent\textbf{$\bullet$ Effect of the Architecture.}
To evaluate the effectiveness of the AMoE-LoRA architecture, we compare it with baseline methods such as LoRA \cite{hu2022lora} and LoRAMoE \cite{dou2023loramoe}. The experiments are designed to verify how AMoE-LoRA improves performance by unifying anomaly-aware and generalist experts adaptation. As shown in Table \ref{ablation_architecture}, AMoE-LoRA achieve superior results across all metrics, which comprehensively demonstrates the significant advantages of our proposed architecture.

\noindent\textbf{$\bullet$ Effect of the Model Parameters.}
To investigate the impact of parameter configurations on the performance of MAU-GPT, we conduct a series of ablation studies. We keep the total number of parameters constant while varying the number of experts in the MoE 22odule and the rank of each expert's LoRA matrix. Specifically, we maintain their product at 64. The results, as shown in Figure \ref{fig:AMoE-LoRA_result} (a), indicate that the model achieves optimal performance on evaluation metrics such as ROUGE and BLEU when the number of experts is set to 16 and the rank of each expert is 4. 

\begin{table}[t]
  \centering
  \resizebox{0.48\textwidth}{!}{
  \begin{tabular}{l|cccc}
    \toprule
    \rowcolor{black!10}\textbf{Architecture} & \textbf{ROUGE-1} & \textbf{ROUGE-2} & \textbf{ROUGE-L} & \textbf{BLEU-4} \\
    \midrule
    LoRA & 0.6971 & 0.4446 & 0.6449 & 0.4529 \\
    LoRAMoE & 0.6991 & 0.4452 & 0.6468 & 0.4598 \\
    \rowcolor[HTML]{E9F3FE}\textbf{AMoE-LoRA} & \textbf{0.7026} & \textbf{0.4537} & \textbf{0.6531} & \textbf{0.4773} \\
    \bottomrule
  \end{tabular}
  }
  \caption{Ablation study about architecture design decisions.}
  \label{ablation_architecture}
\end{table}

\noindent\textbf{$\bullet$ Expert Evaluation.}
We conduct an expert evaluation to further investigate the performance of our model in \ourdataset{}. We randomly sampled 500 QA pairs from Open-Ended QA. Subsequently, we recruit 10 domain experts to perform a blind ranking of the responses generated by LLaVA-1.5-7B, UIO2-Xl-3B, Gemma-3, Yi-VL, InternVL-2.5, AnomalyGPT and MAU-GPT-4B. As shown in Figure \ref{fig:AMoE-LoRA_result} (b), the evaluation results indicate that the responses generated by MAU-GPT are significantly superior to those from other methods in terms of quality and relevance, better meeting real-world application requirements.

\noindent\textbf{$\bullet$ Adaptive Behavior of the Anomaly-Aware Expert.} To examine the adaptability of our anomaly-aware expert, we analyze the LoRA parameters generated by the hypernetwork across different samples and visualize them with t-SNE in 2D. As shown in Figure \ref{fig:AMoE-LoRA_result} (c), the embeddings form clear clusters aligned with object categories, indicating that the hypernetwork produces input-conditioned adapters. Unlike conventional LoRA-based MLLMs with fixed parameters, our model dynamically generates sample-specific LoRA weights, better capturing diverse and subtle industrial anomalies.

\section{Conclusion}
In this work, we present \ourdataset{}, a comprehensive dataset for multi-type industrial anomaly understanding, featuring broad coverage, fine-grained supervision, and a hierarchical QA task structure that spans from binary classification to complex reasoning. To further advance this domain, we introduce MAU-GPT, a domain-adaptive multimodal large model empowered by a novel AMoE-LoRA mechanism that unifies anomaly-aware and generalist experts adaptation. We believe our dataset and model will foster further research in industrial visual understanding and anomaly reasoning.

\section{Acknowledgments}
This work was supported by the National Key Research and Development Program of China (2024YFB3312900), the NSFC (No. 62402426, 62272411, 62441617, 62506333, 62276230), the Key Research and Development Projects in Zhejiang Province (No. 2025C01128, 2024C01106, 2025C01030, 2025C02156), Ningbo Yongjiang Talent Introduction Programme (2023A-400-G), Zhejiang University Education Foundation Qizhen Scholar Foundation, Zhejiang Provincial Natural Science Foundation of China (No. LD25F020001, LDT23F02023F02), the Postdoctoral Fellowship Program of CPSF (GZC20251077), Zhejiang Province Postdoctoral Research Excellence Funding Project (ZJ2025065) and Fundamental Research Funds for the Central Universities (226-2025-00057).

\bibliography{main}

@inproceedings{liu2024improved,
  title={Improved baselines with visual instruction tuning},
  author={Liu, Haotian and Li, Chunyuan and Li, Yuheng and Lee, Yong Jae},
  booktitle={Proceedings of the IEEE/CVF conference on computer vision and pattern recognition},
  pages={26296--26306},
  year={2024}
}

@article{hu2022lora,
  title={Lora: Low-rank adaptation of large language models.},
  author={Hu, Edward J and Shen, Yelong and Wallis, Phillip and Allen-Zhu, Zeyuan and Li, Yuanzhi and Wang, Shean and Wang, Lu and Chen, Weizhu and others},
  journal={ICLR},
  volume={1},
  number={2},
  pages={3},
  year={2022}
}

@article{dou2023loramoe,
  title={LoRAMoE: Alleviate world knowledge forgetting in large language models via MoE-style plugin},
  author={Dou, Shihan and Zhou, Enyu and Liu, Yan and Gao, Songyang and Zhao, Jun and Shen, Wei and Zhou, Yuhao and Xi, Zhiheng and Wang, Xiao and Fan, Xiaoran and others},
  journal={arXiv preprint arXiv:2312.09979},
  year={2023}
}

@inproceedings{li2023blip,
  title={Blip-2: Bootstrapping language-image pre-training with frozen image encoders and large language models},
  author={Li, Junnan and Li, Dongxu and Savarese, Silvio and Hoi, Steven},
  booktitle={International conference on machine learning},
  pages={19730--19742},
  year={2023},
  organization={PMLR}
}

@article{alayrac2022flamingo,
  title={Flamingo: a visual language model for few-shot learning},
  author={Alayrac, Jean-Baptiste and Donahue, Jeff and Luc, Pauline and Miech, Antoine and Barr, Iain and Hasson, Yana and Lenc, Karel and Mensch, Arthur and Millican, Katherine and Reynolds, Malcolm and others},
  journal={Advances in neural information processing systems},
  volume={35},
  pages={23716--23736},
  year={2022}
}

@article{saadany2021bleu,
  title={BLEU, METEOR, BERTScore: evaluation of metrics performance in assessing critical translation errors in sentiment-oriented text},
  author={Saadany, Hadeel and Orasan, Constantin},
  journal={arXiv preprint arXiv:2109.14250},
  year={2021}
}

@article{ganesan2018rouge,
  title={Rouge 2.0: Updated and improved measures for evaluation of summarization tasks},
  author={Ganesan, Kavita},
  journal={arXiv preprint arXiv:1803.01937},
  year={2018}
}

@article{liu2024visual,
  title={Visual instruction tuning},
  author={Liu, Haotian and Li, Chunyuan and Wu, Qingyang and Lee, Yong Jae},
  journal={Advances in neural information processing systems},
  volume={36},
  year={2024}
}

@article{chen2021finqa,
  title={Finqa: A dataset of numerical reasoning over financial data},
  author={Chen, Zhiyu and Chen, Wenhu and Smiley, Charese and Shah, Sameena and Borova, Iana and Langdon, Dylan and Moussa, Reema and Beane, Matt and Huang, Ting-Hao and Routledge, Bryan and others},
  journal={arXiv preprint arXiv:2109.00122},
  year={2021}
}

@article{lu2022learn,
  title={Learn to explain: Multimodal reasoning via thought chains for science question answering},
  author={Lu, Pan and Mishra, Swaroop and Xia, Tanglin and Qiu, Liang and Chang, Kai-Wei and Zhu, Song-Chun and Tafjord, Oyvind and Clark, Peter and Kalyan, Ashwin},
  journal={Advances in Neural Information Processing Systems},
  volume={35},
  pages={2507--2521},
  year={2022}
}

@article{alzarooni2025anomaly,
  title={Anomaly Detection for Industrial Applications, Its Challenges, Solutions, and Future Directions: A Review},
  author={Alzarooni, Abdelrahman and Iqbal, Ehtesham and Khan, Samee Ullah and Javed, Sajid and Moyo, Brain and Abdulrahman, Yusra},
  journal={arXiv preprint arXiv:2501.11310},
  year={2025}
}

@article{ren2022state,
  title={State of the art in defect detection based on machine vision},
  author={Ren, Zhonghe and Fang, Fengzhou and Yan, Ning and Wu, You},
  journal={International Journal of Precision Engineering and Manufacturing-Green Technology},
  volume={9},
  number={2},
  pages={661--691},
  year={2022},
  publisher={Springer}
}

@inproceedings{tang2022industrial,
  title={Industrial Defect Detection Through Computer Vision: A Survey},
  author={Tang, Yunjie and Sun, Kai and Zhao, Danhuai and Lu, Yan and Jiang, Jiaju and Chen, Hong},
  booktitle={2022 7th IEEE International Conference on Data Science in Cyberspace (DSC)},
  pages={605--610},
  year={2022},
  organization={IEEE}
}

@article{bhatt2021image,
  title={Image-based surface defect detection using deep learning: A review},
  author={Bhatt, Prahar M and Malhan, Rishi K and Rajendran, Pradeep and Shah, Brual C and Thakar, Shantanu and Yoon, Yeo Jung and Gupta, Satyandra K},
  journal={Journal of Computing and Information Science in Engineering},
  volume={21},
  number={4},
  pages={040801},
  year={2021},
  publisher={American Society of Mechanical Engineers}
}

@inproceedings{baitieva2024supervised,
  title={Supervised Anomaly Detection for Complex Industrial Images},
  author={Baitieva, Aimira and Hurych, David and Besnier, Victor and Bernard, Olivier},
  booktitle={Proceedings of the IEEE/CVF Conference on Computer Vision and Pattern Recognition},
  pages={17754--17762},
  year={2024}
}

@article{cao2024survey,
  title={A survey on visual anomaly detection: Challenge, approach, and prospect},
  author={Cao, Yunkang and Xu, Xiaohao and Zhang, Jiangning and Cheng, Yuqi and Huang, Xiaonan and Pang, Guansong and Shen, Weiming},
  journal={arXiv preprint arXiv:2401.16402},
  year={2024}
}

@article{Jiang2024mmad,
  title={Mmad: The first-ever comprehensive benchmark for multimodal large language models in industrial anomaly detection},
  author={Jiang, Xi and Li, Jian and Deng, Hanqiu and Liu, Yong and Gao, Bin-Bin and Zhou, Yifeng and Li, Jialin and Wang, Chengjie and Zheng, Feng},
  journal={arXiv preprint arXiv:2410.09453},
  year={2024}
}

@article{tax2004support,
  title={Support vector data description},
  author={Tax, David MJ and Duin, Robert PW},
  journal={Machine learning},
  volume={54},
  pages={45--66},
  year={2004},
  publisher={Springer}
}

@inproceedings{yi2020patch,
  title={Patch svdd: Patch-level svdd for anomaly detection and segmentation},
  author={Yi, Jihun and Yoon, Sungroh},
  booktitle={Proceedings of the Asian conference on computer vision},
  year={2020}
}

@article{bergmann2018improving,
  title={Improving unsupervised defect segmentation by applying structural similarity to autoencoders},
  author={Bergmann, Paul and L{\"o}we, Sindy and Fauser, Michael and Sattlegger, David and Steger, Carsten},
  journal={arXiv preprint arXiv:1807.02011},
  year={2018}
}

@inproceedings{yan2021learning,
  title={Learning semantic context from normal samples for unsupervised anomaly detection},
  author={Yan, Xudong and Zhang, Huaidong and Xu, Xuemiao and Hu, Xiaowei and Heng, Pheng-Ann},
  booktitle={Proceedings of the AAAI conference on artificial intelligence},
  pages={3110--3118},
  year={2021}
}

@inproceedings{mishra2021vt,
  title={VT-ADL: A vision transformer network for image anomaly detection and localization},
  author={Mishra, Pankaj and Verk, Riccardo and Fornasier, Daniele and Piciarelli, Claudio and Foresti, Gian Luca},
  booktitle={2021 IEEE 30th International Symposium on Industrial Electronics (ISIE)},
  pages={01--06},
  year={2021},
  organization={IEEE}
}

@inproceedings{wyatt2022anoddpm,
  title={Anoddpm: Anomaly detection with denoising diffusion probabilistic models using simplex noise},
  author={Wyatt, Julian and Leach, Adam and Schmon, Sebastian M and Willcocks, Chris G},
  booktitle={Proceedings of the IEEE/CVF Conference on Computer Vision and Pattern Recognition},
  pages={650--656},
  year={2022}
}

@inproceedings{chu2020neural,
  title={Neural batch sampling with reinforcement learning for semi-supervised anomaly detection},
  author={Chu, Wen-Hsuan and Kitani, Kris M},
  booktitle={Computer Vision--ECCV 2020: 16th European Conference, Glasgow, UK, August 23--28, 2020, Proceedings, Part XXVI 16},
  pages={751--766},
  year={2020},
  organization={Springer}
}

@article{liznerski2020explainable,
  title={Explainable deep one-class classification},
  author={Liznerski, Philipp and Ruff, Lukas and Vandermeulen, Robert A and Franks, Billy Joe and Kloft, Marius and M{\"u}ller, Klaus-Robert},
  journal={arXiv preprint arXiv:2007.01760},
  year={2020}
}

@article{bovzivc2021mixed,
  title={Mixed supervision for surface-defect detection: From weakly to fully supervised learning},
  author={Bo{\v{z}}i{\v{c}}, Jakob and Tabernik, Domen and Sko{\v{c}}aj, Danijel},
  journal={Computers in Industry},
  volume={129},
  pages={103459},
  year={2021},
  publisher={Elsevier}
}

@article{pang2021explainable,
  title={Explainable deep few-shot anomaly detection with deviation networks},
  author={Pang, Guansong and Ding, Choubo and Shen, Chunhua and Hengel, Anton van den},
  journal={arXiv preprint arXiv:2108.00462},
  year={2021}
}

@inproceedings{ding2022catching,
  title={Catching both gray and black swans: Open-set supervised anomaly detection},
  author={Ding, Choubo and Pang, Guansong and Shen, Chunhua},
  booktitle={Proceedings of the IEEE/CVF conference on computer vision and pattern recognition},
  pages={7388--7398},
  year={2022}
}

@article{huang2020surface,
  title={Surface defect saliency of magnetic tile},
  author={Huang, Yibin and Qiu, Congying and Yuan, Kui},
  journal={The Visual Computer},
  volume={36},
  number={1},
  pages={85--96},
  year={2020},
  publisher={Springer}
}

@article{tang2019online,
  title={Online PCB defect detector on a new PCB defect dataset},
  author={Tang, Sanli and He, Fan and Huang, Xiaolin and Yang, Jie},
  journal={arXiv preprint arXiv:1902.06197},
  year={2019}
}

@inproceedings{jezek2021deep,
  title={Deep learning-based defect detection of metal parts: evaluating current methods in complex conditions},
  author={Jezek, Stepan and Jonak, Martin and Burget, Radim and Dvorak, Pavel and Skotak, Milos},
  booktitle={2021 13th International congress on ultra modern telecommunications and control systems and workshops (ICUMT)},
  pages={66--71},
  year={2021},
  organization={IEEE}
}

@inproceedings{bergmann2019mvtec,
  title={MVTec AD--A comprehensive real-world dataset for unsupervised anomaly detection},
  author={Bergmann, Paul and Fauser, Michael and Sattlegger, David and Steger, Carsten},
  booktitle={Proceedings of the IEEE/CVF conference on computer vision and pattern recognition},
  pages={9592--9600},
  year={2019}
}

@inproceedings{wieler2007weakly,
  title={Weakly supervised learning for industrial optical inspection},
  author={Wieler, Matthias and Hahn, Tobias},
  booktitle={DAGM symposium in},
  volume={6},
  pages={11},
  year={2007}
}

@article{chen2024expanding,
  title={Expanding performance boundaries of open-source multimodal models with model, data, and test-time scaling},
  author={Chen, Zhe and Wang, Weiyun and Cao, Yue and Liu, Yangzhou and Gao, Zhangwei and Cui, Erfei and Zhu, Jinguo and Ye, Shenglong and Tian, Hao and Liu, Zhaoyang and others},
  journal={arXiv preprint arXiv:2412.05271},
  year={2024}
}

@article{moryossef2019step,
  title={Step-by-step: Separating planning from realization in neural data-to-text generation},
  author={Moryossef, Amit and Goldberg, Yoav and Dagan, Ido},
  journal={arXiv preprint arXiv:1904.03396},
  year={2019}
}

@inproceedings{bolukbasi2017adaptive,
  title={Adaptive neural networks for efficient inference},
  author={Bolukbasi, Tolga and Wang, Joseph and Dekel, Ofer and Saligrama, Venkatesh},
  booktitle={ICML},
  year={2017}
}

@inproceedings{li2022devil,
  title={The devil is in the labels: Noisy label correction for robust scene graph generation},
  author={Li, Lin and Chen, Long and Huang, Yifeng and Zhang, Zhimeng and Zhang, Songyang and Xiao, Jun},
  booktitle={Proceedings of the IEEE/CVF Conference on Computer Vision and Pattern Recognition},
  pages={18869--18878},
  year={2022}
}

@article{li2023zero,
  title={Zero-shot visual relation detection via composite visual cues from large language models},
  author={Li, Lin and Xiao, Jun and Chen, Guikun and Shao, Jian and Zhuang, Yueting and Chen, Long},
  journal={Advances in Neural Information Processing Systems},
  volume={36},
  pages={50105--50116},
  year={2023}
}

@article{shao2024knowledge,
  title={Knowledge-guided causal intervention for weakly-supervised object localization},
  author={Shao, Feifei and Luo, Yawei and Gao, Fei and Yang, Yi and Xiao, Jun},
  journal={IEEE Transactions on Knowledge and Data Engineering},
  volume={36},
  number={11},
  pages={6477--6489},
  year={2024},
  publisher={IEEE}
}

@inproceedings{shao2021improving,
  title={Improving weakly supervised object localization via causal intervention},
  author={Shao, Feifei and Luo, Yawei and Zhang, Li and Ye, Lu and Tang, Siliang and Yang, Yi and Xiao, Jun},
  booktitle={Proceedings of the 29th ACM International Conference on Multimedia},
  pages={3321--3329},
  year={2021}
}

\end{document}